\documentclass[letterpaper, 10 pt, conference]{ieeeconf}  % Comment this line out if you need a4paper

\IEEEoverridecommandlockouts                              % This command is only needed if
\usepackage{filecontents}

\usepackage{graphicx} % for pdf, bitmapped graphics files
\usepackage{mathptmx} % assumes new font selection scheme installed
\usepackage{times} % assumes new font selection scheme installed
\usepackage{amsmath} % assumes amsmath package installed
\usepackage{amssymb}  % assumes amsmath package installed
\usepackage{float}
\usepackage{floatflt}
\usepackage[font=small]{caption}
\usepackage{subfigure}
\usepackage[nolist,nohyperlinks]{acronym}

\usepackage[backend=biber,style=ieee,doi=true,isbn=false,url=false,eprint=true,clearlang=true,]{biblatex}
\usepackage{hyperref}
\PassOptionsToPackage{breaklinks=true}{hyperref}
\usepackage[breaklinks=true]{hyperref}
\usepackage{bbold}

\usepackage{array}
\usepackage{booktabs}
\DeclareMathOperator*{\argmax}{argmax}

\title{\LARGE \bf
Modular Sensor Fusion for Semantic Segmentation
}

\author{Hermann Blum, Abel Gawel, Roland Siegwart and Cesar Cadena\\% <-this % stops a space
\thanks{Authors are with the Autonomous Systems Lab of ETH Zurich.
{\tt\small \{blumh, gawela, rsiegwart, cesarc\}@ethz.ch}}%
\thanks{
This research has received funding from the EU H2020 research project under grant agreement No 688652, the Swiss State Secretariat for Education, Research and Innovation (SERI) No 15.0284, and by the Swiss National Science Foundation through the National Center of Competence in Research Robotics (NCCR).}
}

\begin{document}
\begin{minipage}{\textwidth}
\copyright 2018 IEEE. Personal use of this material is permitted. Permission from IEEE must be obtained for all
other uses, in any current or future media, including reprinting/republishing this material for advertising
or promotional purposes, creating new collective works, for resale or redistribution to servers or lists, or
reuse of any copyrighted component of this work in other works.\\

Please cite this paper as:\\%
\begin{verbatim}
@inproceedings{blum2018fusion,
  title     = "Modular Sensor Fusion for Semantic Segmentation",
  author    = "Blum, Hermann and Gawel, Abel and Siegwart, Roland and Cadena, Cesar",
  booktitle = "2018 {IEEE/RSJ} International Conference on Intelligent Robots
               and Systems ({IROS})",
  year      = 2018;
}
\end{verbatim}

\end{minipage}

\maketitle

%%%%%%%%%%%%%%%%%%%%%%%%%%%%%%%%%%%%%%%%%%%%%%%%%%%%%%%%%%%%%%%%%%%%%%%%%%%%%%%%
\begin{abstract}
Sensor fusion is a fundamental process in robotic systems as it extends the perceptual range and increases robustness in real-world operations.
Current multi-sensor deep learning based semantic segmentation approaches do not provide robustness to under-performing classes in one modality, or require a specific architecture with access to the full aligned multi-sensor training data. 
In this work, we analyze statistical fusion approaches for semantic segmentation that overcome these drawbacks while keeping a competitive performance.
The studied approaches are modular by construction, allowing to have different training sets per modality and only a much smaller subset is needed to calibrate the statistical models.
We evaluate a range of statistical fusion approaches and report their performance against state-of-the-art baselines on both real-world and simulated data.
In our experiments, the approach improves performance in IoU over the best single modality segmentation results by up to 5\%.
We make all implementations and configurations publicly available.

\end{abstract}

\section{Introduction}%
\label{sec:intro}
Semantic segmentation has become a popular discipline in recent years~\cite{garcia2017areview}.
It most commonly deals with the pixel-wise categorical classification of image data, but can be employed for various sensor data, e.g., 3D data~\cite{charles2017pointnet, wu20153dshapenet}.
In robotics, semantic segmentation is relevant for scene understanding in autonomous driving~\cite{ess2009segmentation}, localization tasks~\cite{gawel2018xview}, or natural human-machine interaction~\cite{oberweger2015hands}.
While architectures for single modalities, e.g., RGB or 3D data are becoming increasingly accurate, there remain perceptually difficult cases that single sensors are unable to reliably classify.
Lane-markings and Pictures on a wall are invisible to depth sensors. 
RGB cameras, on the other hand, are much more sensitive to weather and lighting conditions.
Using multiple sensors can increase performance, compensating for other sensors' weaknesses or failures.
Recently, several approaches were proposed to address the challenge of fusing multiple sensor inputs for semantic segmentation.
One avenue of research leads towards training one specific network for all modalities together~\cite{Hazirbas2016-ti,Xiang2017-sv,Cadena2016-tq}.
Another avenue is to leverage single-modality \emph{expert} networks and a trained fusion network~\cite{Mees2016-ks, Valada2017-hr, ros2016synthia}.
One major pitfall of these solutions is the requirement of training for any sensor combinations and therewith aligned multi-modal training data.

\begin{figure}
    \centering
    \includegraphics[width=0.6\linewidth]{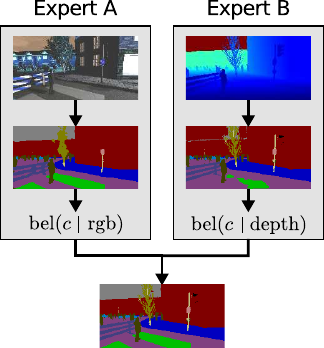}
    \caption{Individual semantic segmentation classifiers are combined modularly using different statistical methods, resulting in improved semantic segmentation without additional training.}
    \vspace{-5mm}
\end{figure}

In contrast, we wish to design segmentation systems that remain modular, and can fuse different \emph{expert} networks independently trained on single modalities.
Apart from neural networks, fusion of classifers has been a part of Machine Learning literature for decades already~\cite{Xu1992-cn}.
Dependent on well-known statistics like confusion matrices, classifier outputs are combined using statistical methods.

In this work, we propose a novel, scalable semantic segmentation fusion architecture based on separately trained expert networks that are statistically fused using Bayesian fusion or Dirichlet Fusion.
The individual \emph{experts} can thus be trained on different datasets, and no additional training is required to fuse their outputs.
In addition, due to its modularity, input modalities can be added and removed on-the-fly. 
Hence, failures of single \emph{experts} can be compensated if detected.

This paper presents the following contribution:
\begin{itemize}
\item A novel statistical fusion method for multi-modal semantic segmentation based on the Dirichlet distribution.
\item The first fusion network alternative that does not require training on aligned data.
\item Evaluation and analysis of different statistical fusion methods on real-world and synthetic datasets.
\end{itemize}

\section{Related Work}

\subsection{Semantic Segmentation}%
\label{rel:sec:segmentation}

Semantic Segmentation is commonly understood as the task of pixel-wise classification of image data~\cite{garcia2017areview}.
While being an important discipline in the Computer Vision community, it has relevance for various robotics applications, especially scene understanding and interaction~\cite{ess2009segmentation, gawel2018xview, oberweger2015hands}.
Presently, the best performing algorithms for semantic segmentation on popular benchmarks, e.g., PascalVOC~\cite{Everingham2010-ci} are based on \acp{CNN}.
Here, a common architecture is the encoder-decoder structure~\cite{Noh2015-kx,Badrinarayanan2017-rd,Long2015-fm}.
The encoder aims to capture features on multiple abstractions of the images, as commonly used in image classification tasks.
The decoder then unfolds the feature data back to the pixel-level.

In this work, we base our semantic segmentation structure on the \ac{FCN}~\cite{Long2015-fm} and AdapNet~\cite{Valada2017-hr}.

\subsection{Fusion}%
\label{rel:sec:fusion}
While semantic segmentation is a maturing discipline for RGB-based systems, a fusion with complementary sensor data can further improve the segmentation results~\cite{Hazirbas2016-ti,Valada2017-hr}, especially in challenging classes for the visual sensor.
Based on \ac{CNN} architectures, different mechanisms for fusion have been developed.
In principle, we distinguish between two architectures, i.e., networks fully designed for fusion~\cite{Hazirbas2016-ti,Xiang2017-sv,Cadena2016-tq}, and unimodal \emph{expert} networks with an added fusion network~\cite{Mees2016-ks,Valada2017-hr}, as originally proposed by Jacobs et al.~\cite{Jacobs1991}.
FuseNet~\cite{Hazirbas2016-ti} fuses features from RGB and Depth images gathered from two VGG16-encoders~\cite{vgg16} and decodes the fused features into a semantic segmentation.
%
%\hbcomm{I am not sure whether we should mention the DA-RNN here as it does not really test it's fusion mechanism.}
A similar approach is used in~\cite{Xiang2017-sv}, where features are extracted from different layers' outputs of the encoder in an adapted \ac{FCN} structure.
A multimodal autoencoder is proposed in~\cite{Cadena2016-tq} where the possible deficiencies in one modality have to be foreseen and introduced in training time.
The works by \cite{Valada2017-hr} and \cite{Mees2016-ks} explore different mechanisms that fuse classifier outputs at a later stage, they find that a gating network learning factors for a weighted sum of the individual segmentation outputs works best for their network.

In contrast, our approach is inspired by work on fusion of classifiers~\cite{Xu1992-cn}.
While the original work deals with uni-modal handwriting recognition, we extend it towards multi-modal semantic segmentation by replacing the simple classification output with a full output score vector.
In addition, we omit the possibility of rejecting classifications present in the original work.

Further general fusion techniques have been developed to deal with ensembles of neural networks.
These include averaging over a range of experts or voting principles.
In such a setup, classifiers are trained equally to specialize on different aspects of a problem, e.g., by employing techniques like bagging~\cite{Breiman1996-dy} and boosting~\cite{Schapire1990-dh}.
However, as we deal with architectures that train on different input modalities, we do not have the same control over the specific strengths and weaknesses of the different classifiers,  which is necessary for simple fusion techniques like voting to work.
Moreover, these techniques require a multitude of expert networks, which is impracticable with big \acp{CNN}.

\section{Method}

\subsection{Semantic Segmentation}

As baseline systems for semantic segmentation we use two different neural network architectures.
%an \ac{FCN}-based and a ResNet-based architecture.
%
The \ac{FCN} structure was first introduced by Long et al.\ in 2015~\cite{Long2015-fm} and used by Xiang et al. to fuse different modalities~\cite{Xiang2017-sv}. 
The \ac{FCN} uses the VGG-16 encoder~\cite{vgg16}, which consists of iterating \(3\times3\) convolutions and \(2\times2\) max-pooling layers that maps an input image onto a lower dimensional feature map.
This feature map is then scaled up again with deconvolutions and mapped onto the output class using \(1\times1\) convolutions.
When we refer to the \ac{FCN} in experiments, we mean the version shown in Figure~\ref{fig:fcn} as used in~\cite{Xiang2017-sv}, which reduces complexity by replacing the trainable deconvolutions with simple bilinear interpolation. 
While not achieving the performance of more complex networks, we used this simpler model for testing many different fusion methods and report the comparison.

AdapNet~\cite{Valada2017-hr} is a structure designed to improve some of the shortcomings of the \ac{FCN}.
It uses the ResNet encoder~\cite{resnet} where every dimensionality-reducing stage is split into convolutions applied on different scales of the input to make the feature map more independent against different scales of an input object.
Similar to the \ac{FCN} above, different scales of this feature-map are then processed independently and stacked together before using \(1\times1\) convolutions and deconvolutions to map the features onto the output classes. 
While the training is slower and complex, Valada et al. found the evaluation time of the AdapNet to be faster than that of the \ac{FCN} and other methods~\cite{Valada2017-hr}.

\subsection{Statistical Classifier Combination}%
\label{method:fusion}

For modality fusion, we train individual baseline networks for every input modality completely independent of each other.
The fusion is then applied based on the outputs of all these different baselines that were evaluated on the same scene.

All of the approaches introduced in~\ref{rel:sec:fusion} have in common that the fusion process is part of the network structure, i.e. the fusion is not adaptable to different sensor combinations, and requires retraining. 
Moreover, with the exception of the gating network \cite{Valada2017-hr}, none of these systems can deal with crashing sensors by exchanging or disabling input experts.

As a more modular approach, we propose a fusion technique that is based on statistically merging the outputs of individual classifiers. 
In this sense, this work is heavily inspired by techniques described by Xu et al.\ in 1992~\cite{Xu1992-cn}.

\textbf{Bayes Categorical Fusion}:
For every pixel we want to produce a probability $p(k|\textrm{all expert outputs})$ for every possible class $k \in {1, ..., K}$, given the outputs of all uni-modal experts. 
From this, we can then choose for every pixel the class with the highest probability.
This is sometimes called the believe of a class $\textrm{bel}(k)$. 
Based on Xu et al.~\cite{Xu1992-cn} and Bayes' formula, we find:

\begin{align*}
    p(k|\textrm{all expert outputs}) &\propto p(\textrm{all expert outputs} |k) \, p(k)\\
    &\propto p(k) \prod \limits_{i\in\textrm{modalities}} p(\textrm{out}_i|k)
\end{align*}
with $\textrm{out}_i$ the classification output of expert $i$.
$p(\textrm{out}_i|k)$ is a categorical distribution over the expert's classification output, which we will know at inference time.
We use that the conditionals $p(\textrm{out}_i|k)$ are independent of each other, as the interdependency between the different modalities is exactly given by the ground truth class $k$. 
\begin{figure}
    \centering
    \includegraphics[width=0.6\linewidth]{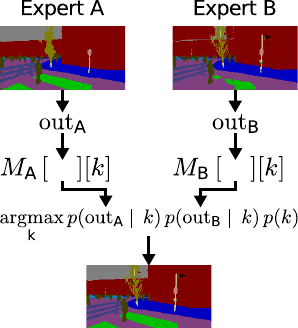}
    \caption{Example of the Bayesian categorical fusion with 2 modalities. The classification output is used as index to the confusion matrix in order to produce the conditional class likelihoods. For fusion, the class with the biggest joint likelihood is choosen.}
    \label{fig:bayes_fusion}
    \vspace{-5mm}
\end{figure}

Each of the individual conditional categorical distributions is given by the confusion matrix $M_i$ of the corresponding classifier.
If the first dimension of the matrix is the actual expert output $\textrm{out}_i$ and the second dimension is the ground-truth class $k$, it follows:
\begin{align*}
    p(\textrm{out}_i|k) &= \frac{M_i\lbrack\textrm{out}_i\rbrack\lbrack k\rbrack}{\sum_{j=1}^{K} M_i\lbrack j\rbrack \lbrack k \rbrack}
\end{align*}
The prior $p(k)$ can also be set on basis of the class-occurrence in these confusion matrices.

We find the fused classification by choosing the class that has the highest log-probability:
\begin{align}
    \textrm{output class} &= \argmax \limits_k p(k|\textrm{all expert outputs}) \nonumber\\
    &= \argmax \limits_k \left\lbrack \log p(k) + \sum \limits_{i\in\textrm{experts}} \log p(\textrm{out}_i|k) \right\rbrack \label{fusion_decision}
\end{align}

A schematic of this technique is shown in Figure~\ref{fig:bayes_fusion}.

\textbf{Dirichlet Fusion}:
In the above approach, we fuse the different experts on basis of their classification output.
However, this disregards important information, especially in such cases of special interest for fusion where two or more classes are equally likely.
To take these cases into account, we have to produce $p(\pmb{y}|k)$ that is dependent on the full softmax output vector $\pmb{y}$ and therefore contains score values for all possible classes. 
This distribution is given by the Dirichlet distribution, the conjugate prior of the categorical distribution.
\begin{align*}
    \pmb{y} &\sim \textrm{Dirichlet}(\pmb{\alpha})\\
    \textrm{pdf}(\pmb{y}) &= \frac{\Gamma\left(\sum_{j=1}^{K} \alpha_j \right)}{\prod_{j=1}^{K} \Gamma(\alpha_j)} \prod \limits_{j=1}^{K} y_j^{\alpha_j - 1}
\end{align*}
The concentration parameters $\pmb{\alpha}$ of this distribution are constrained by $\alpha_j > 0 \; \forall \alpha_j \in \pmb{\alpha}$. The higher $\alpha_l / \sum_{j=1}^K \alpha_j$, the higher is the probability of a vector close to $y_j = \mathbb{1}_{j=l}$.

In order to find the correct concentration parameters for every conditional class and every expert, we make use of the fast \ac{EM} algorithm developed by Minka~\cite{Minka2000-ll} and improved and translated to python by Sklar~\cite{Sklar2014-lx}.
Both make use of the fact that there is a sufficient statistic for the \ac{EM} fitting of the Dirichlet distribution $\mathbb{s} = \frac{1}{N}\sum_{n=1}^N \log \pmb{y}^{(n)}$.
Without the sufficient statistic over a set of $N$ pixels, the M-Step would have to compute over all the images in every iteration.
Instead, $\mathbb{s}$ can be produced from measurements over the data before starting the \ac{EM} algorithm, which then can validate the likelihood of the produced parameters at every iteration against the sufficient statistic.
The standard loss function of the \ac{EM} algorithm is the likelihood of the parameters $\pmb{\alpha}$ against the data:
\begin{align}
    \mathcal{L}(\pmb{\alpha}, \mathbb{s}) =& N \left\lbrack \log \Gamma \left( \sum \limits_j \alpha_j \right) - \sum \limits_j \log \Gamma (\alpha_j) \right. \nonumber \\
    &+ \left. \sum\limits_j (\alpha_j - 1)\log \mathbb{s}_j \right\rbrack \label{standard_loss}\\
    \mathbb{s}_j =& \frac{1}{N}\sum_{n=1}^{N} \log y_j^{(n)} \nonumber
\end{align}
We found that this standard method does produce reasonable parameters \(\pmb{\alpha}\), but the resulting conditional likelihoods do not work well in our decision function from Equation~(\ref{fusion_decision}).
The main problems are edge cases where two classes get similar scores in \(\pmb{y}\).
In this case, conditionals found with Equation~(\ref{standard_loss}) assign equally low likelihoods to all classes. 
Therefore, we introduced 2 regularization terms:
We added the norm \(\sum \alpha_j^2\) as a regularization term to prevent the concentration parameters to grow too large, which in turn would produce low likelihoods for any \(\pmb{y}\) that is not close to a one-hot vector.
In addition, we do not aim for parameters \(\pmb{\alpha} \) that explain our classifier output best, but we want to distinguish between different conditionals.
Therefore, we build a sufficient statistic \(\bar{\mathbb{s}}\) of all classifier outputs for different ground-truth than the conditional class we search for and add \(-\beta \,  \mathcal{L}(\pmb{\alpha}, \bar{\mathbb{s}})\) into the Loss.
With these 2 additions, and omitting the constant factor \(N\), as proposed by Sklar~\cite{Sklar2014-lx}, we arrive at the following loss function for the \ac{EM} algorithm:
\begin{align*}
    \mathcal{L}'(\alpha, \mathbb{s}, \bar{\mathbb{s}}) =& (1 - \beta) \left\lbrack\log \Gamma \left( \sum \limits_j \alpha_j \right) - \sum \limits_j \log \Gamma (\alpha_j) \right\rbrack \\
    &- \beta \sum\limits_j (\alpha_j - 1)\log \bar{\mathbb{s}}\\
    &+ \sum\limits_j (\alpha_j - 1)\log \mathbb{s} - \delta \sum\limits_j \alpha_j^2
\end{align*}

Once the concentration parameters for all ground-truth classes \(\pmb{\alpha}^{(k)}\) are found, fusion can be performed with the following decision function:
\begin{align*}
    \textrm{output class} &= \argmax \limits_k \left\lbrack \log p(k) +  \sum \limits_{i\in\textrm{experts}} \log f(\pmb{y}_i|k) \right\rbrack\\
    f(\pmb{y}_i|k) &= \textrm{pdf}(\pmb{y}_i, \pmb{\alpha}_i^{(k)})
\end{align*}

\section{Experiments} % (fold)
\label{experiments}
We conduct experiments with implementations in tensorflow%
%~\cite{tensorflow2015-whitepaper}
and train with RMS Prop~\cite{rmsprop} using default configurations.

We test our methods on the Synthia-Rand-Cityscapes~\cite{ros2016synthia} and the Cityscapes datasets~\cite{Cordts2016Cityscapes}, both showing urban street scenes.
The Synthia dataset comes from simulation and features alongside RGB also very precise depth images, and pixel-wise semantic segmentation into 13 classes.
The real-world Cityscapes dataset contains RGB, noisy disparity images from stereo matching, and pixel-wise semantic segmentation into 30 different classes. 
For both datasets, we use a common set of 12 classes that was also used in~\cite{Valada2017-hr}: \emph{void, sky, building, road, sidewalk, fence, vegetation, pole, car/bus/truck, traffic sign, pedestrian, bicycle/motorcycle/rider}.
Furthermore, we resized input images to 768x384 following the experiments of~\cite{Valada2017-hr}.
As there is no given split between train- and test-set in the Synthia-Rand-Cityscapes, we take a random 10\% sample as test-set, and another 10\% sample as a validation and development set. The images in the dataset are produced from random viewpoints, which makes this simple split feasible.
The development set is used to compute the confusion matrices and conditional Dirichlet distributions.
For Cityscapes, we take a random 5\% sample out of the given training set as development set.
The parameters $\beta$ and $\delta$ for the Dirichlet fusion are found with a grid search on the development set before evaluating the method with the found parameters on the separate test set.
We run the \ac{EM} algorithm with a maximum of 1000 iterations.

During training, we employ cropping and flipping as augmentation methods, after which we finetune the baselines on small batches of full images.
The semantic segmentation performance is always measured in \ac{IoU}~\cite{Everingham2010-ci}. 
For overall performance, we take the mean over all available classes.
In cases where we do not report the per-class \ac{IoU}, we additionally report \ac{AP}.

All implementations and configurations of the experiments are available at \url{https://github.com/ethz-asl/modular_semantic_segmentation}.

\subsection{Fusion on Synthetic Data}
\label{exp:syncity}
\begin{figure}
    \centering
    \includegraphics[width=\linewidth]{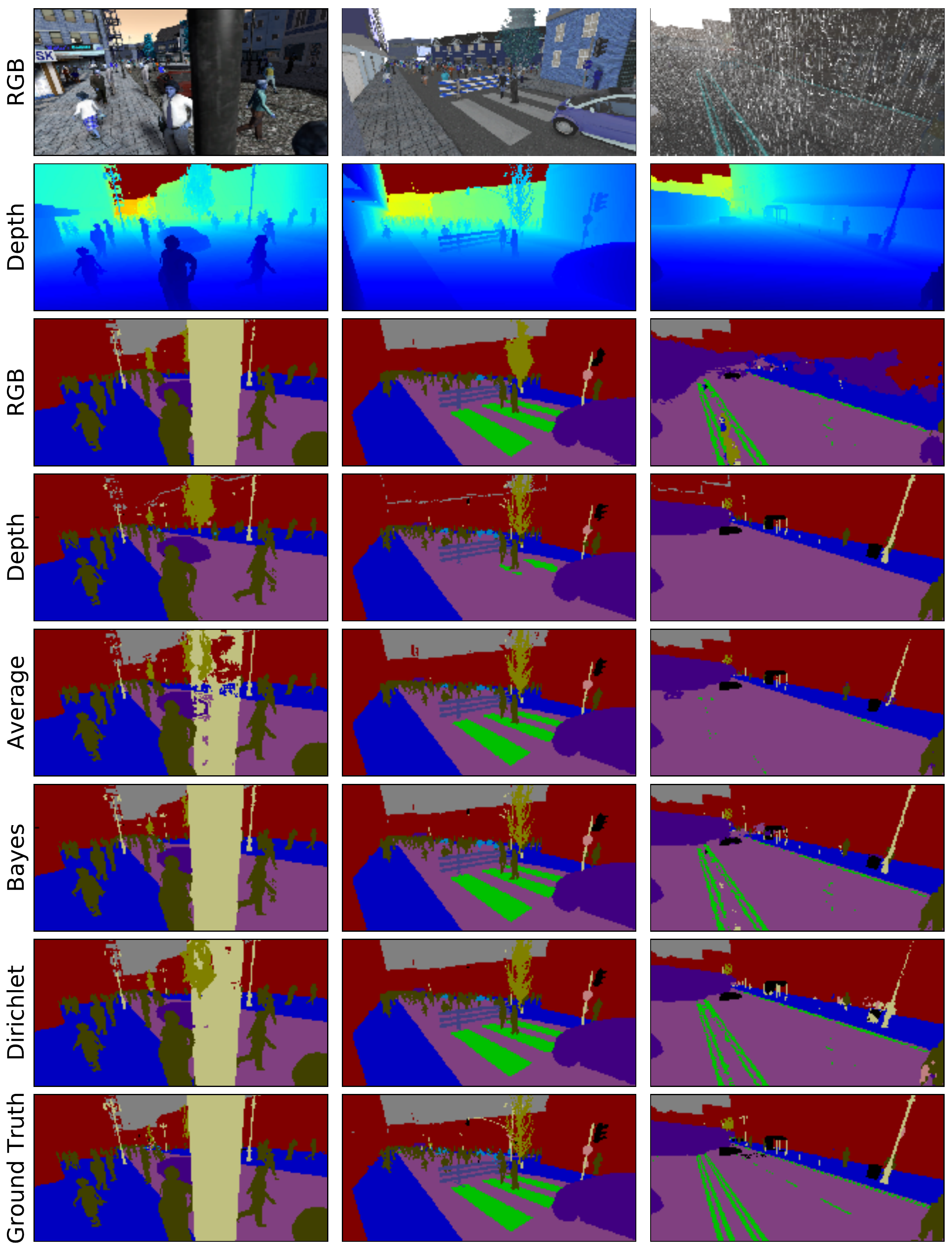}
    \caption{Qualitative Examples from the Synthia Rand Cityscapes Fusion Experiments. \textbf{Left} Due to some error in the simulation, the pole right in Front of the camera is not visible to the depth expert. Because the class-scores from the RGB experts are very high, the fusion methods mostly recover from this failure. 
    \textbf{Center} Lanemarkings are not visible to the depth expert, even tough it predicts them below pedestrians crossing the street. The correct classification from the RGB expert is fused into the output by all methods.
    \textbf{Right} Fusion can improve robustness for different weather conditions, as shown here for heavy rain occluding most of the RGB input.}
    \label{fig:synthia_examples}
    \vspace{-7mm}
\end{figure}

We design 2 experimental setups.
First, we use the previously listed 12 classes and later we add lane-markings as an additional class.
In this context, lane-markings are a very interesting example for fusion as they are indistinguishable from roads on a depth image and therefore only visible to an RGB expert.
Following the evaluation of~\cite{Valada2017-hr} for semantic segmentation of RGB and Depth images, we compare Averaging, and our Bayes Categorical and Dirichlet fusion%
%\footnote{Unfortunately we could neither reproduce the results reported by~\cite{Valada2017-hr} nor get access to their baselines, nor fusion results, of any of their reported experiments. Therefore, a fair benchmark against their method was not possible.}
.
Averaging fuses the experts by taking the mean over all softmax outputs and choosing the class with the highest mean score.
Tables~\ref{tab:fusion:syncity} and~\ref{tab:synthia:details} show the results of our evaluation, Figure~\ref{fig:synthia_examples} shows qualitative examples.
%

\iffalse
\begin{table}
    \caption{Fusion of AdapNet Baselines on Synthia Rand Cityscapes}
    \centering
    \label{tab:fusion:syncity}
    \begin{tabular}{rcccc}
            \toprule
        Method & \multicolumn{2}{c}{\parbox[b]{2cm}{\centering without\\ Lane-markings}} & \multicolumn{2}{c}{\parbox[b]{2cm}{\centering with\\ Lane-markings}}\\

        & IoU & AP & IoU & AP\\
        \midrule
              RGB  &  73.39  &  80.87  &  75.92  &  83.34  \\
            Depth  &  72.70  &  79.54  &  63.76  &  70.20  \\
          Average  &  \textbf{78.70}  &  84.04  &  79.05  &  83.65  \\
            Bayes  &  78.62  &  \textbf{86.12}  &  79.91  &  86.92  \\
        Dirichlet  &  77.27  &  83.31  &  \textbf{80.19}  &  \textbf{87.43}  \\
        \bottomrule
    \end{tabular}
\end{table}
\fi

\begin{table}
    \caption{Fusion of AdapNet Baselines on Synthia Rand Cityscapes}
    \vspace{-3mm}
    \centering
    \label{tab:fusion:syncity}
    \begin{tabular}{llccccc}
        \toprule
        \parbox[b]{1cm}{Lane-\\markings} & & Dirichlet & Bayes & Average & RGB & Depth\\
        \midrule
        no & IoU  & 77.27 & 78.62 & \textbf{78.70} & 73.39 & 72.70\\
        no & AP   & 83.31 & \textbf{86.12} & 84.04 & 80.87 & 79.54\\
        \midrule
        yes & IoU & \textbf{80.19} & 79.91 & 79.05 & 75.92 & 63.76\\
        yes & AP  & \textbf{87.43} & 86.92 & 83.65 & 83.34 & 70.20\\
        \bottomrule
    \end{tabular}
    \vspace{-1mm}
\end{table}

\begin{table}
    \caption{Per-Class IoU of AdapNet Baselines on Synthia Rand Cityscapes with Lanemarkings}
    \vspace{-3mm}
    \centering
    \label{tab:synthia:details}
    \begin{tabular}{rccccc}
        \toprule
        Class & Dirichlet & Bayes &  Average  &  RGB  & Depth\\
\midrule
          Mean  &  \textbf{80.19}  &  79.91  &  79.05  &  75.92  &  63.76 \\
\midrule
           Sky  &  97.39  &  \textbf{97.41}  &  95.57  &  95.54  &  9.47 \\
      Building  &  \textbf{96.85}  &  96.76  &  96.20  &  93.70  &  79.86 \\
          Road  &  \textbf{94.09}  &  92.78  &  93.11  &  91.58  &  88.92 \\
      Sidewalk  &  \textbf{95.07}  &  94.04  &  94.70  &  91.71  &  93.60 \\
         Fence  &  \textbf{74.76}  &  72.97  &  74.54  &  68.30  &  71.89 \\
    Vegetation  &  89.89  &  \textbf{90.62}  &  90.02  &  81.42  &  90.32 \\
          Pole  &  65.53  &  \textbf{66.42}  &  64.79  &  56.17  &  60.19 \\
           Car  &  \textbf{93.19}  &  91.39  &  92.81  &  87.81  &  90.56 \\
  Traffic Sign  &  41.75  &  \textbf{51.54}  &  48.31  &  45.77  &  37.76 \\
    Pedestrian  &  \textbf{79.94}  &  74.73  &  79.72  &  73.70  &  72.77 \\
       Bicycle  &  \textbf{65.47}  &  63.11  &  64.42  &  58.57  &  59.40 \\
   Lane-marking  &  \textbf{68.29}  &  67.17  &  54.39  &  66.73  &  10.44 \\

        \bottomrule
    \end{tabular}
    \vspace{-7mm}
\end{table}

\begin{table}
    \caption{Fusion of AdapNet Baselines on Cityscapes}
    \vspace{-3mm}
    \centering
    \label{tab:cityscapes}
    \begin{tabular}{rccccc}
\toprule
          Class & Dirichlet  &  Bayes  &  Average  &  RGB  &  Depth\\
\midrule
          Mean  &  \textbf{69.22}  &  68.77  &  68.47  &  69.20  &  54.12 \\
\midrule
           Sky  &  86.97  &  90.45  &  88.93  &  \textbf{90.54}  &  78.60 \\
      Building  &  \textbf{84.60}  &  83.59  &  84.32  &  84.09  &  72.83 \\
          Road  &  92.37  &  91.71  &  \textbf{92.59}  &  91.53  &  91.79 \\
      Sidewalk  &  \textbf{67.66}  &  65.01  &  67.25  &  66.42  &  57.58 \\
         Fence  &  \textbf{41.54}  &  37.92  &  40.79  &  39.84  &  23.62 \\
    Vegetation  &  87.13  &  \textbf{88.04}  &  84.48  &  \textbf{88.04}  &  66.56 \\
          Pole  &  \textbf{44.23}  &  42.11  &  43.77  &  41.94  &  33.28 \\
       Vehicle  &  \textbf{87.62}  &  86.20  &  87.02  &  86.54  &  77.92 \\
  Traffic Sign  &  51.29  &  51.37  &  48.33  &  \textbf{52.12}  &  18.85 \\
        Person  &  \textbf{64.90}  &  63.54  &  64.75  &  63.54  &  52.27 \\
       Bicycle  &  53.13  &  56.54  &  50.98  &  \textbf{56.57}  &  22.00 \\
\bottomrule
    \end{tabular}
    \vspace{-5mm}
\end{table}

In general, we find that the statistical fusion significantly improves the semantic segmentation with respect to the two uni-modal baselines.
We also find that the inclusion of lane-markings changes the comparison between the different fusion methods.
Without lane-markings, we observe only a minor difference between averaging and the Bayes Categorical fusion.
However, when including lane-markings, we see that the training of the Depth expert is much harder, resulting in a worse individual performance of this modality alone.

One effect of this is that the Bayes categorical fusion improves over the averaging due to its ability to `pick' the better performing expert based on the classification outputs and mirror its performance. %

The second effect is that the Dirichlet fusion is producing the best results in most individual classes, and the overall mean performance. %
We observe that in the classes where the Dirichlet fusion produces the best results, it usually also improves the classification with respect to both individual experts, instead of mirroring the performance of the better one. %
We offer the conclusion that this is possible through the more detailed input into the fusion mechanism.

\subsection{Fusion on Real-World Data} % (fold)
\label{exp:cityscapes}

\begin{figure}
    \centering
    \includegraphics[width=\linewidth]{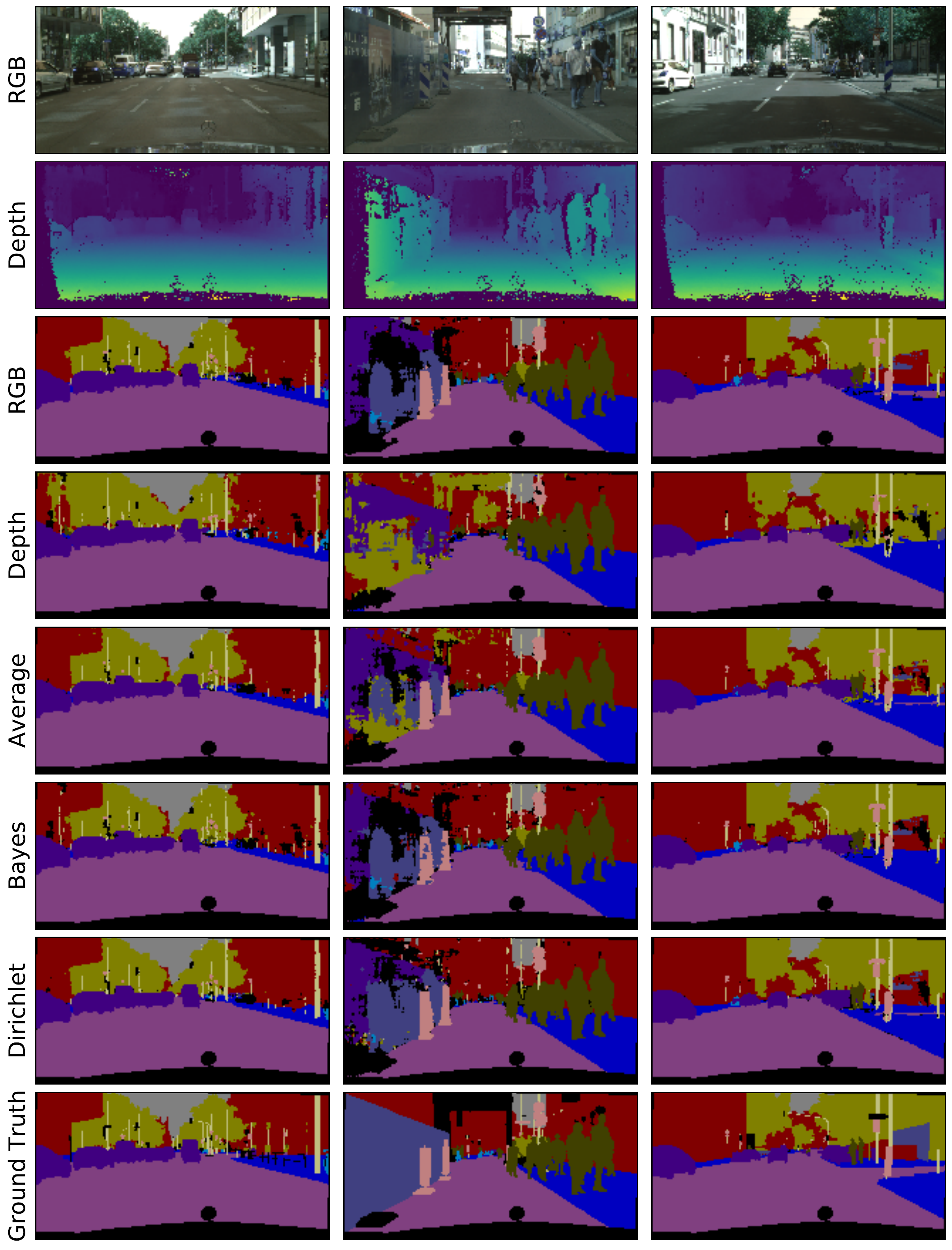}
    \caption{Qualitative Examples from the Cityscapes Fusion Experiments. As the depth images are visibly worse than in the synthetic data, and there are no different weather conditions, fusion improvements are less obvious. The biggest improvements can be found for poles in the background, as well as the left wall in the center column.}
    \label{fig:cityscapes_examples}
    \vspace{-7mm}
\end{figure}

To validate the results from the synthetic data, we test the same methods on the Cityscapes~\cite{Cordts2016Cityscapes} dataset. The results of this experiment are shown in table~\ref{tab:cityscapes}. Qualitative examples can be found in figure~\ref{fig:cityscapes_examples}. 
The Synthia Rand Cityscapes is especially designed to cover the same classes and urban street scenes present in Cityscapes, enabling a fair comparison of the results.
The Cityscapes dataset does not provide depth images from a separate depth sensors, but disparity maps computed from stereo cameras.
Due to considerable noise in the depth estimation, we observe in general much worse performance of the Depth expert, compared to the experiments on synthetic data.
We also find that the fusion methods offer no significant improvement in segmentation, compared to the RGB expert.
While the fusion improves the segmentation on classes such as \emph{pole} or \emph{vehicle}, it has lower performance than the RGB expert on classes like \emph{sky}. %

\begin{figure*}[tb]
    \centering
    \includegraphics[width=\linewidth]{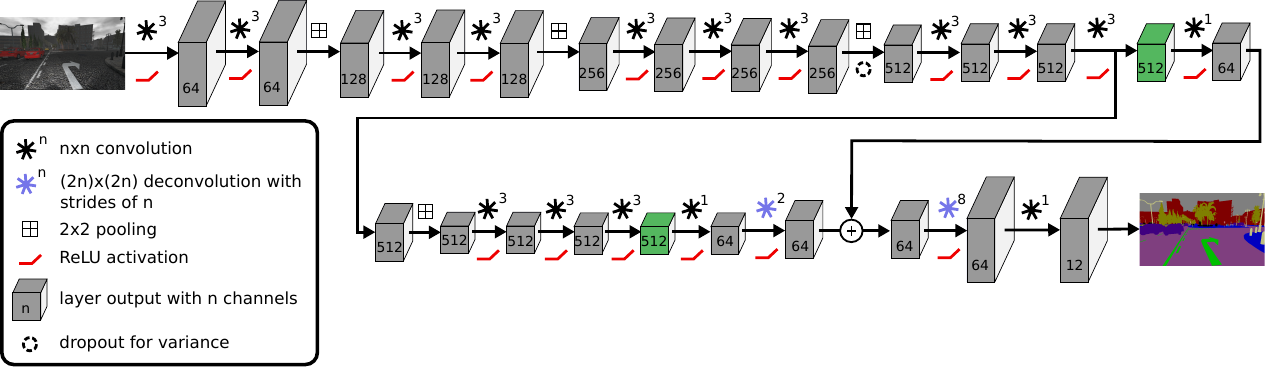}
    \caption{The \ac{FCN} architecture adapted from ~\cite{Xiang2017-sv} and used in our experiments. It consists of the VGG-16 encoder~\cite{vgg16} of stacked convolutions, and poolings. Instead of trainable deconvolutions, this architecture uses bilinear interpolations for the decoder, thus reducing the complexity of the network. Dropout is applied only for the variance fusion experiment. In case of multiple encoders, the marked green encoder outputs are stacked together before applying the $1\times 1$ convolution.}
    \label{fig:fcn}
    \vspace{-5mm}
\end{figure*}

\subsection{Modalities Trained on Different Datasets}

An important advantage of our modular fusion methods is that unimodal experts do not require simultaneous training.
They can be trained independently, even on different datasets, as not every modality may be available for every dataset.

We therefore test our fusion methods in the following scenario: There is very little labelled data for a new modality (depth) available.
We take a Depth baseline that is trained on simulation data and just plug it into our system, calibrating it against a few labelled example images.

In this experiment, we take the Depth baseline trained on synthia data and fuse it together with the RGB baseline from experiment~\ref{exp:cityscapes}, validating the system on the cityscapes test set.
The results are shown in Table~\ref{tab:synthetic_depth}.
While averaging was a competitive method in the earlier experiments, it fails in this experiment as it is the only method without calibrating the experts.
We can also see that the validation of the experts is working as expected.
In fact, the simple Bayes Categorical fusion is mirroring the output of the RGB baseline.

For those classes where the depth expert carries some information, the Dirichlet fusion is always learning to improve the output of the RGB expert.
The overall fusion results for Bayes and Dirichlet fusion are even better than with the depth baseline trained on Cityscapes. 
We attribute this to the observation that the Bayes and Dirichlet fusion perform well in a setting where they can choose between the experts, but as this experiment shows their model does not cope well with one expert that is always worse.

\begin{table}
    \caption{Fusion of AdapNet Baselines on Cityscapes. RGB is trained on Cityscapes, Depth is exclusively trained on Synthia Rand Cityscapes.}
    \label{tab:synthetic_depth}
    \centering
    \begin{tabular}{rccccc}
\toprule
          Class & Dirichlet  &  Bayes  &  Average  &  RGB  &  Depth\\
\midrule
          Mean  &  \textbf{69.28}  &  69.20  &  62.65  &  69.20  &  3.37 \\
\midrule
           Sky  &  90.16  &  90.53  &  89.63  &  \textbf{90.54}  &  0.00 \\
      Building  &  \textbf{84.18}  &  84.09  &  68.99  &  84.09  &  14.09 \\
          Road  &  91.56  &  91.53  &  \textbf{91.63}  &  91.53  &  0.00 \\
      Sidewalk  &  \textbf{66.45}  &  66.42  &  62.91  &  66.42  &  0.03 \\
         Fence  &  39.75  &  \textbf{39.84}  &  35.76  &  \textbf{39.84}  &  0.00 \\
    Vegetation  &  \textbf{88.07}  &  88.04  &  79.85  &  88.04  &  22.60 \\
          Pole  &  \textbf{42.22}  &  41.94  &  32.34  &  41.94  &  0.36 \\
       Vehicle  &  \textbf{86.82}  &  86.54  &  86.15  &  86.54  &  0.00 \\
  Traffic Sign  &  \textbf{52.37}  &  52.11  &  43.21  &  52.12  &  0.00 \\
        Person  &  \textbf{63.89}  &  63.54  &  50.87  &  63.54  &  0.00 \\
       Bicycle  &  \textbf{56.60}  &  56.57  &  47.86  &  56.57  &  0.00 \\
\bottomrule
    \end{tabular}
    \vspace{-5mm}
\end{table}

\subsection{Further Benchmarks}

\begin{table*}
    \caption{\ac{FCN}-based Fusion Methods on Synthia Rand Cityscapes}
    \centering
    \label{tab:fcn_benchmarks}
    \begin{tabular}{lccccccc}
        \toprule
        & FuseFCN & Average & Bayes & Dirichlet & Variance & RGB & Depth\\
        \midrule
        IoU & 76.90 & 76.38 & 74.99 & 66.96 & 66.35 & 72.24 & 72.01\\
        AP  & 83.79 & 83.54 & 82.91 & 73.82 & 73.83 & 80.35 & 81.15\\
        Inference Time
            & $72 \pm 22$ ms % experiment 1059
            & $43 \pm 11$ ms % experiment 1064
            & $46 \pm 16$ ms % experiment 1071
            & $52 \pm 24$ ms % experiment 1067
            & $310 \pm 18$ ms % experiment 1065
            & $22 \pm 11$ ms % experiment 1062
            & $22 \pm 12$ ms % experiment 1063
            \\
        \bottomrule
    \end{tabular}
    \vspace{-5mm}
\end{table*}

In this section, we conduct additional tests with different \ac{FCN}-based~\cite{Long2015-fm} architectures and further investigate possibilities of modular fusion and the performance of our proposed methods.
The principle architecture we use is shown in Figure~\ref{fig:fcn}.

In the experiments before, we could compare the improvements of each method with the uni-modal baselines. As we now compare between fusion methods that are not based on any baselines, this is no longer possible. We therefore train all \ac{FCN}s to their best possible performance and compare overall results.

Additionally, we evaluate the inference time per image of every method. The inference time is measured over 10000 trials and every method is evaluated on constant input of the same size as an image, to further reduce the influence of caching and data loading on the measurements. All computations are conducted on a single GPU. Dependent on the available hardware, the methods may benefit differently from different degrees of parallellisation.

\paragraph{Fusion by Variance}
Using Dropout-Monte-Carlo~\cite{Gal2016-zk}, we can measure model uncertainties from each modality expert at inference time.
In theory, this variance $\sigma_j^2$ should contain information about the uncertainties of the different classifiers, trained on different modalities.
The per-pixel certainty $\omega$ is then approximated by
\begin{align*}
    \omega = 1 / \left(\frac{1}{K} \sum \limits_{j=0}^{K-1} \sigma_j^2 
    \right)
\end{align*}
Where $K$ is the number of classes and $\sigma^2$ is measured from the softmax output class scores $\pmb{y}$ over a number of samples.
We fuse the experts  with a weighted sum.
\begin{align}
    \pmb{y}_\textrm{fused} = \frac{\sum_{i=0}^{M-1} \omega_i \pmb{y}_i}{\sum_{i=0}^{M-1} \omega_i}
\end{align}
The class score vector $\pmb{y}_i$ of every uni-modal expert $0\leq i < M$ is weighted by the certainty of the given pixel and expert.
Note that if the certainties of all the experts are assumed equal, this is reduced to the averaging fusion used before.
During the development we found that this method is very sensitive to the type of dropout performed.
In particular, any dropout close to the output layer leads to very noisy fusion outputs, requiring a large number of Monte-Carlo samples to compensate.
Following the findings from~\cite{Kendall2015-cg}, we choose to introduce a dropout layer after the third pooling, before the network branches into two parts and not immediately before a pooling layer. 
This is also indicated in Figure~\ref{fig:fcn}.

\paragraph{Full Fusion Network}
As described in~\ref{rel:sec:fusion}, fusion is often conducted with a fully integrated network.
As a benchmark we use the structure from~\cite{Xiang2017-sv} with two encoders fused together into one decoder, but without the recurrent layer.
It is therefore following the architecture shown in figure~\ref{fig:fcn}.
We train on aligned RGB and Depth images until convergence.

The results are shown in Table~\ref{tab:fcn_benchmarks}.
The experiment is again conducted with the standard set of 12 classes.

As expected, the fully integrated network expresses best performance.
Contradictory to the experiments with AdapNet on the same dataset reported in section~\ref{exp:syncity}, the averaging fusion is performing significantly better than both the Bayes categorical and the Dirichlet fusion.
We attribute this to the fact that both baselines have very similar performances.

We find that the variance fusion and the Dirichlet fusion express lower performance than the uni-modal experts.
Both methods are based on mechanisms that attempt to measure uncertainties of individual experts.
Further investigation for the Dirichlet fusion revealed that opposed to AdapNet, the \ac{FCN} architecture often produces outputs that assign similar probabilities to more than two different classes.
The AdapNet baselines from section~\ref{exp:syncity}, however, usually express higher certainty and assign high probability scores to one or two classes, also at the border of objects.
We conclude that the Dirichlet fusion method, in particular the proposed \ac{EM} fitting, is not able to capture the variability of the \ac{FCN} output well.

While the Average Fusion is on average faster than the Bayes or Dirichlet fusion, as would be expected from the complexity of the calculation, the difference in inference time between the three methods stays within one standard deviation. We also note that the modular methods are all faster then the integrated FusionFCN, even tough this architecture is a `late-fusion' method with independent encoders.

\section{Discussion}

Our results indicate that statistical fusion of modalities is a promising avenue for semantic segmentation in cases where we cannot or do not want to retrain a fusion network on aligned data.
Despite the good performance of the averaging on the synthetic data, we argue that it is only advisable to use this technique when the experts have comparably good performance.
Especially in real-world applications, we usually have sensors expressing much better performance than the others.
Here, the averaging fails to produce convincing results.

The statistical fusion is able to exploit the information even in cases of very strong performance differences and improve the segmentation result.
Although not expressing superior performance in all tested cases, statistical fusion generally improves the segmentation performance over single modality systems without the need for cumbersome training.

We notice that the Dirichlet model is in general the best performing technique.
However, the current \ac{EM} algorithm to find the concentration parameters suffers under very flat, under-confident class probability outputs of the single experts.
In these cases, the Dirichlet model cannot produce parameters that result in suitable decision functions for the fusion.
Consequently, in cases of under-confident uni-modal experts, the Dirichlet fusion looses its expressiveness power to combine the modalities on every class.

A general finding from our experiments is that the Bayes Categorical fusion, and also the Dirichlet fusion, work best with experts that have complementary strengths.
Here, these frameworks show their power to pick the best performing expert for every class or even learn on basis of their combination.
In practice, we showed that this depends both on the set of classes as well as the quality of the input data for the different modalities.

An analysis of the inference time for the different methods further shows that statistical fusion is a promising method of time-critical systems, as the inference time is significantly lower than a fully integrated fusion network architecture.

\section{Conclusions and Outlook}

In this paper, we have presented and evaluated two modular approaches to fuse multiple modalities for semantic segmentation.
The novel proposed Dirichlet fusion shows the best results of the statistical fusion methods, especially when using modalities with complementary strengths and weaknesses. 
Furthermore, the modularity in terms of the used segmentation experts allows for a seamless extension to new experts without re-training the already existing ones.

The performance of the proposed fusion scheme is close to the performance of specifically trained fusion networks, but requires no additional training on aligned data.
It therefore gives wider access to datasets that do only contain a single modality.

The findings are consistent over different datasets from simulation to real world scenarios.
The biggest problems with the statistical fusion were encountered in cases of low-confidence classifier outputs.
In future work, we will therefore test whether measurements of input based uncertainties of the neural network classifiers can further improve the results of statistical fusion.

To conclude, the proposed statistical fusion promises to be a powerful basis of a modular framework for semantic segmentation.
With this framework, we can produce semantic scene descriptions for a diverse set of robots, enabling collaboration and mutual scene understanding.

\printbibliography

\begin{acronym}
\acro{FCN}{Fully Convolutional Network}
\acro{CNN}{Convolutional Neural Network}
\acro{IoU}{Intersection over Union}
\acro{EM}{Expectation-Maximization}
\acro{AP}{Average Precision}
\end{acronym}
\end{document}